\pdfoutput=1

\documentclass[11pt]{article}
\usepackage{color, colortbl}

\usepackage{siunitx}
\usepackage{arabtex}
\usepackage{utf8}
\setcode{utf8}
\usepackage{tabularx}
\usepackage{commath}
\usepackage{emnlp-2023}

\usepackage{times}
\usepackage{latexsym}

\usepackage[T1]{fontenc}

\usepackage[utf8]{inputenc}

\usepackage{caption}
\usepackage{subcaption}
\usepackage{microtype}

\usepackage{inconsolata}

\usepackage{amsmath,amsfonts,bm}

\def\eqref#1{equation~\ref{#1}}

\def\1{\bm{1}}

\DeclareMathAlphabet{\mathsfit}{\encodingdefault}{\sfdefault}{m}{sl}
\SetMathAlphabet{\mathsfit}{bold}{\encodingdefault}{\sfdefault}{bx}{n}

\usepackage{booktabs}
\usepackage{comment}
\usepackage{hyperref}
\usepackage{url}
\usepackage{algorithm}
\usepackage{algorithmic}
\usepackage{amsmath}
\usepackage{multirow}
\usepackage{graphicx}
\usepackage{xcolor}
\usepackage{pgf}

\usepackage{color, colortbl}
\usepackage{booktabs}

\newcommand{\annote}[3]{{\color{#3}%
		\colorbox{#3}{\bfseries\sffamily\tiny\textcolor{white}{#2}}
		\color{#3}
		\footnotesize\emph{#1}}%
}

\newcommand{\todo}[1]{\annote{#1}{TODO}{red}}
\newcommand{\nllb}{\textsc{nllb}}
\newcommand{\holisticbias}{\textsc{HolisticBias}}
\newcommand{\multilingualholisticbias}{\textsc{MultilingualHolisticBias}}

\usepackage{newfloat}
\usepackage{listings}
\floatstyle{ruled}
\newfloat{listing}{tb}{lst}{}
\floatname{listing}{Listing}
\begin{document}

\title{Multilingual Holistic Bias: Extending Descriptors and Patterns \\to Unveil Demographic Biases in Languages at Scale}
\author{Marta R. Costa-jussà, Pierre Andrews, Eric Smith, Prangthip Hansanti,\\ \textbf{Christophe Ropers, Elahe Kalbassi, Cynthia Gao, Daniel Licht, Carleigh Wood}\\
$\dag$Meta AI \\
\texttt{\{costajussa,mortimer,ems,prangthiphansanti,}\\
\texttt{chrisropers,ekalbassi,cynthiagao,dlicht,carleighwood\}@meta.com}}

\newcommand{\fix}{\marginpar{FIX}}
\newcommand{\new}{\marginpar{NEW}}

\floatstyle{ruled}

\maketitle
\begin{abstract}
We introduce a multilingual extension of the \holisticbias{} dataset, the largest English template-based taxonomy of textual people references: \multilingualholisticbias. This extension consists of 20,459 sentences in 50 languages distributed across all 13 demographic axes. Source sentences are built from combinations of 118 demographic descriptors and three patterns, excluding nonsensical combinations. Multilingual translations include alternatives for gendered languages that cover gendered translations when there is ambiguity in English.
Our benchmark is intended to uncover demographic imbalances and be the tool to quantify mitigations towards them.  

Our initial findings show that translation quality for EN-to-XX translations is an average of 8 spBLEU better when evaluating with the masculine human reference compared to feminine. In the opposite direction, XX-to-EN, we compare the robustness of the model when the source input only differs in gender (masculine or feminine) and masculine translations are an average of almost 4 spBLEU better than feminine. When embedding sentences to a joint multilingual sentence representations space, we find that for most languages masculine translations are significantly closer to the English neutral sentences when embedded.

\end{abstract}


\section{Introduction}
\label{sec:introduction}


Demographic biases are relatively infrequent phenomena but present a very important problem. The development of datasets in this area has raised the interest in evaluating Natural Language Processing (NLP) models beyond standard quality terms. This can be illustrated by the fact that machine translation (MT) models systematically translate neutral source sentences into masculine or feminine depending on the stereotypical usage of the word (e.g. ``homemakers'' into “amas de casa”, which is the feminine form in Spanish and ``doctors'' into “médicos”, which is the masculine form in Spanish). While gender is one aspect of demographic biases, we can further explore abilities, nationalities, races or religion and observe other generalizations of the models that may perpetuate or amplify stereotypes and inequalities in society. Quantifying and evaluating these biases is not straightforward because of the lack of datasets and evaluation metrics in this direction. Proper evaluation will enable further mitigation of these biases.

\paragraph{Related work} \holisticbias{} \cite{smith-etal-2022-im} is an English dataset built from templated sentences that can elicit enough examples in various contexts to analyze and draw actionable conclusions: when measuring toxicity after translating \holisticbias{} prompts \cite{costajussa2023toxicity}; when measuring the relative perplexity of different sentences as a function of gendered noun or descriptor \cite{smith-etal-2022-im}; when looking at skews of the usages of different descriptors in the training data, etc. Other datasets consisting of slotting terms into templates were introduced by \cite{kurita2019measuring,may2019measuring,sheng2019woman,brown2020language,webster2020measuring}, to name a few. The advantage of templates is that terms can be swapped in and out to measure different forms of social biases, such as stereotypical associations. Other strategies for creating bias datasets include careful handcrafting of grammars \cite{renduchintala-williams-2022-investigating}, collecting prompts from the beginnings of existing text sentences \cite{dhamala2021bold}, and swapping demographic terms in existing text, either heuristically \cite{papakipos2022augly} or using trained neural language models \cite{qian2022perturbation}. Most of these alternatives cover few languages or they are limited in the bias scope (e.g. only gender \cite{stanovsky-etal-2019-evaluating,renduchintala-etal-2021-gender,levy-etal-2021-collecting-large,costajussaetal:2022,renduchintala-williams-2022-investigating}), which forbids the evaluation of highly multilingual MT systems.

\paragraph{Contributions} Our work approaches this problem by carefully translating a subset of the \holisticbias{} dataset into 50 languages (see appendix A for a complete list), covering 13 demographic axes. As an extension of \holisticbias{}, we will invite additions and amendments to the dataset, in order to contribute to its establishment as a standardized method for evaluating bias for highly multilingual NLP models. We use the proposed dataset to experiment on MT and sentence representation.
Results when translating from English show an average 8 spBLEU reduction when evaluating on the feminine reference set compared to masculine. This showcases the preference towards masculine translations. Among the 13 demographic axes of \holisticbias{}, the quality of translation averaged across languages is highest for the nationality axis and lowest for the cultural axis. Results when translating to English show that the masculine set has almost 4 spBLEU improvement compared to the feminine set. When embedding sentences to a joint multilingual sentence representations space 
which is the core tool of multilingual data mining, we find that for most languages, there is a significant 
difference in the similarity between the masculine translations and the feminine one. Masculine translations are significantly closer to the English sentence when embedded, even if this difference remains small and we do not yet know the effect on the mining algorithm.

\section{Background: \holisticbias{} dataset}
\holisticbias{} is composed of 26 templates, more than 600 descriptors (covering 13 demographic axes) and 30 nouns.  Overall, this dataset consists of over 472k English sentences  used in the context of a two-person conversation. Sentences are typically created from combining a sentence template (e.g., “I am a [NOUN PHRASE].”), a noun (e.g., parent), and a descriptor (e.g., disabled). The list of nearly 600 descriptors covers 13 demographic axes such as ability, race/ethnicity, or gender/sex. The noun can imply a certain gender (e.g. woman, man) or avoid gender references (e.g. child, kid). Sentence templates allow for both singular and plural forms of the descriptor/noun phrase.

Experiments in MT with the NLLB model using the full initial (English only) version of the \holisticbias{} dataset, as reported in \cite{costajussa2023toxicity}, show that the percentage of true added toxicity is also relatively low (from 0.004\% in Chinese to 1.07\% in Kinyarwanda) but that the number of examples in absolute value is much greater (20 in Chinese, 4,951 in Kinyarwanda) due to the fact that \holisticbias{} is composed of roughly 230 times more sentences than the FLORES-200 dev and devtest sets put together. The templated nature of \holisticbias{} also makes it possible to observe different translation behaviors for the same lexical items in different contexts.

Even if it is ideal for prompting English language models and MT from English to other languages, the main shortcomings of the \holisticbias{} dataset are that we cannot evaluate how the quality varies for this particular domain; and we cannot study biases in a variety of languages, which affects multilingual NLP applications. 

\holisticbias{} successfully combines sentence patterns, descriptors and nouns. However, when considering its extension to multiple languages, we did not consider a translation of patterns, descriptors and nouns because languages don't have the same, universal lexical, syntactic or morphological schemes, and not all languages allow for the use of templating/slotting like was done for English. Instead, we did a full translation of selected sentences in \holisticbias{} as described in the next section.

\section{\multilingualholisticbias{} Dataset}

\holisticbias{} is a large dataset and it is unfeasible to translate sentence-by-sentence to a large multilingual set of languages. In this section, we describe the data selection criteria, then the translation guidelines and finally we report the statistics of the final \multilingualholisticbias{} dataset.

\subsection{\holisticbias{} sentence selection} 

We selected a total of 118 descriptors and three templates. Descriptor terms were selected based on the  44 most toxic ones in NLLB translation \cite{costajussa2023toxicity}, 45 descriptors that are not robust in translation (excluding descriptors with hyphens due to their complexity), and 29 random ones. These descriptors cover 13 demographic axes as shown in Table \ref{tab:descriptors}.

\begin{table*}[h!]
\centering
\scriptsize
\begin{tabular}{|l|l|}  
\hline
Axis & Descriptors\\
\hline
Ability  & aphasic, disabled, dyslexic, handicapped, hearing-impaired, low-vision, mute, neurodiverse (reworded to “neurodivergent”), \\
& neurotypical, paralyzed, quadriplegic, who incurred a traumatic brain injury, who is an amputee, who is deaf, who is \\
& visually impaired, who walks with crutches, with a cognitive disability, with a developmental disability, with a disability,\\
& with a lisp, with a stutter, with an intellectual disability \\
\hline
Age &40-year-old, 50-year-old, 55-year-old, adolescent, nonagenarian, octogenarian, teenage, thirty-something \\
\hline
Body type & balding, barrel-chested, black-haired, brunette, burly, chubby, chunky, curly-haired, dark-haired, dark-skinned, four-eyed,\\
& gangly, ginger, goateed, graying (reworded to “going gray”), kinky-haired, mildly overweight, morbidly obese, muscular,\\
& mustachioed, overweight, plump, ponytailed, scruffy, slightly overweight, toned, underweight \\
\hline
Characteristics & American-born, addict, alcoholic, homemaker, in the Reserve, non-native (reworded to “immigrant”), orphan, part-timer,\\ 
& refugee, undocumented, widowed, with a criminal background, with a gambling problem, with dementia, with no kids\\
\hline
Cultural & Boomer, Millennial, hillbilly\\
\hline
Gender and sex & agender, androgynous, cisgender, gender non-conforming, genderfluid, genderqueer, hermaphrodite, intersex, trans,\\& trans femme, trans masc, transsexual\\
\hline
Gender and sex & queer\\ and Sexual orientation & \\
\hline
Nationality & Dominican, Salvadoran\\
\hline
Race and ethnicity & Aboriginal, Asian, Black, Hispanic, Indigenous, Native America\\
\hline
Religion & Bahá'í, Confucianist, Evangelical, Hasidic, Mormon, Rastafarian, Shintoist, Sikh, Spiritualist, Unitarian, Wiccan, \\ & Zoroastrian\\
\hline
Sexual orientation &
asexual, bisexual, demisexual, gay, heterosexual, pansexual, polyamorous\\
\hline
Socioeconomic class & one-percenter, wealthy, with a master's degree\\
\hline

  \end{tabular}
  \caption{List of complete descriptors classified by demographic axes for \multilingualholisticbias{}. \label{tab:descriptors}}
\end{table*}

We use three templates that refer to people in 3 different ways, people, person and children: ``I have friends who are [descriptor] people.'', ``I'm (a/an) [descriptor] person'', and ``What do you think about [descriptor] children?''
We prioritize neutral nouns referring to people because these tend to generate stereotypical translations when translating into gendered languages like Romance ones. We choose a mixture of singular and plural.  
We manually excluded any descriptors that: (1) do not make sense with child(ren), eg.  "I'm a senior-citizen child", "I'm an unmarried child", replacing the noun with "veteran(s)"; (2) focus on beauty/ugliness because of being demographically uninteresting eg. "dirty-blonde",
(3) have a tendency to be always pejorative ("trailer trash"); (4) are US-specific (“Cuban-American”); (5) are English specific (e.g. “blicket”, a purposefully nonsense term following English phonological rules); (6) are relatively rare (“affianced’); (7) overlap with another term in the dataset (“American Indian” vs. “Native American”).

\subsection{Translation Guidelines}


The objective of our dataset is to have gold-standard human translations from professional linguists that are accurately faithful to the source sentences. 
The additional challenge of the bias data set is that the source sentences generated via the templated approach are vague and disconnected from any disambiguating context. Therefore, linguists needed to make sure that their translations were both accurate enough to not include bias and generic enough as to be used in most possible contexts. Linguists were asked to:

\begin{enumerate}
    \item provide accurate and culturally appropriate translations;
    \item provide separate translations for each noun class or grammatical gender for languages that make use of noun classes or grammatical genders;
    \item avoid relying on their personal experience to translate (especially descriptors), given that personal experience is where bias may exist; instead, conduct lexical research through credible sources of information, such as unilingual dictionaries or encyclopedias, and provide information as to the source being used and the rationale for selecting one translation over another;
    \item remain faithful to the source (see below for further details on faithfulness to the source). 
\end{enumerate}
  
Being faithful to the source is a north-star principle in any translation exercise, which can sometimes conflict with other guidance frequently given to translators, such as the need to produce fluent or natural-sounding translations. The two principles are complementary when the source material contains naturally produced, fluent language. However, due to the templated nature of the source material in our particular case, some source sentences may appear lacking in fluency (especially when using the nouns \textit{people} or \textit{person}). The question therefore arose whether these nouns should be translated or omitted. The general guidance given to linguists was that (1) they should bear in mind that the source sentences may not necessarily sound fluent or natural to native speakers of the source language (here, English) and they should strive to remain faithful to the source, and (2) they should feel free to omit such nouns if they feel that the resulting translation sounds unacceptable in their respective native languages.

Additionally, we established a peer-review process, which involved reviewers from different vendors. This added an extra layer of quality checks to ensure accuracy and consistency in the final translation. This process was similar to translation quality checks in which two reviewers provided by different vendors are assigned to work together to review, refute the translation from the translating vendor, and suggest the most appropriate one, if necessary. All research and discussions by reviewers were documented to ensure transparency and accountability. This crucial step helped us track the changes made to the original translation and identify issues that may arise during the translation process. The reviewed translation is considered the final one. 


\subsection{Data Statistics}
\label{sec:datastatistics}
\begin{figure*}
    \includegraphics[width=\textwidth]{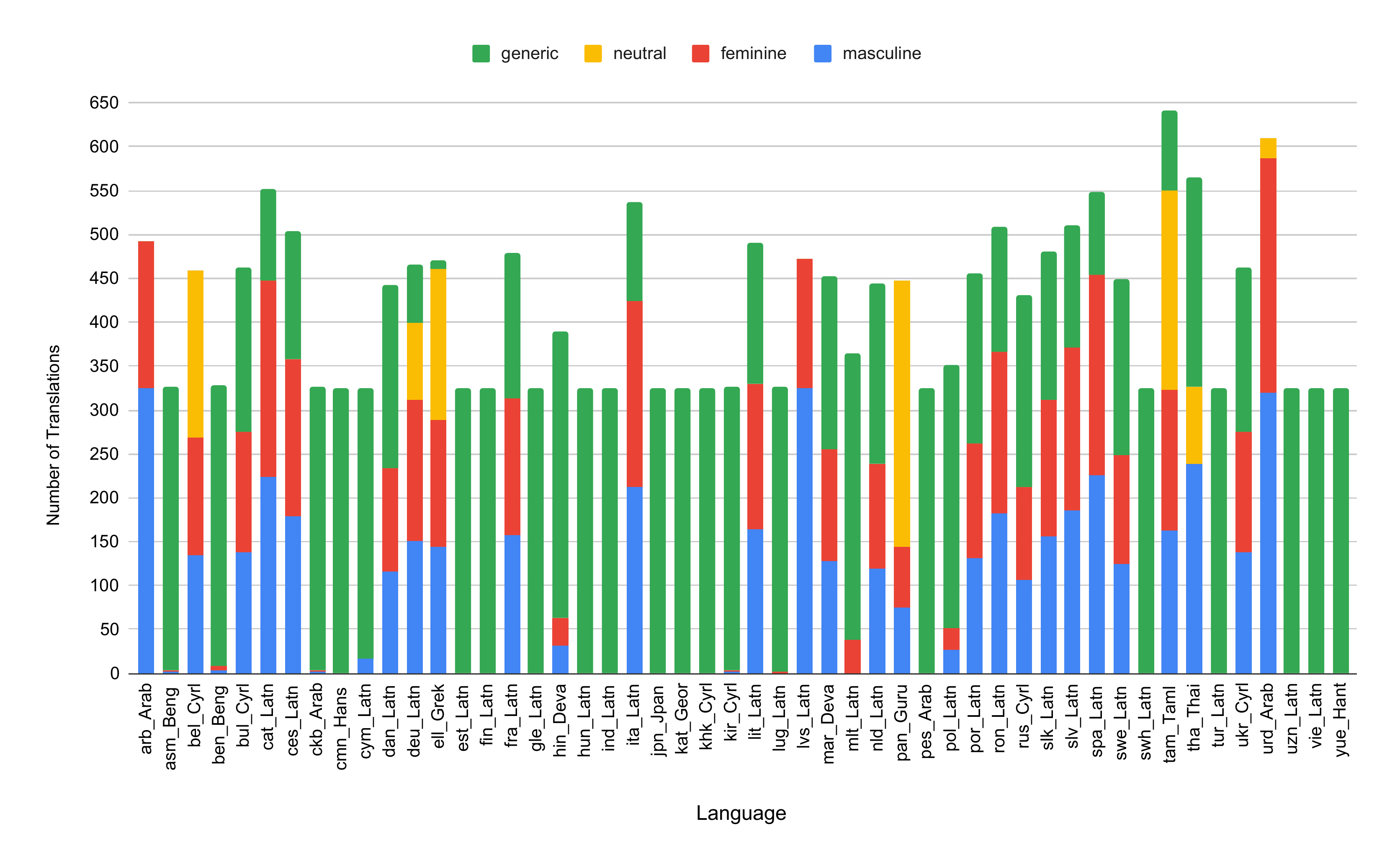}
    \caption{Number of human translations per language and gender (masculine, feminine, neutral and generic).}
    \label{fig:twotrans}
\end{figure*}

Altogether, our initial English dataset consists of 325 sentences.
Figure \ref{fig:twotrans} shows the number of translations for each gender (masculine, feminine, neutral and generic). 
There are 15 languages\footnote{Chinese (simplified), Estonian, Finish, Irish, Hungarian, Indonesian, Japanese, Georgian, Halh Mongolian, Persian, Swahili, Turkish, Northen Uzbeck, Vietnamese, Yue Chinese (traditional)} 
for which we only have the generic human translation. Those languages do not show feminine and masculine inflections for the patterns that we have chosen. Among the other languages where have several translations, the number of sentences for each gender varies.
For the languages in which we have gender inflections, \multilingualholisticbias{} keeps separated sets: one for each gender representation (masculine, feminine, neutral and generic). 


\section{Machine Translation Evaluation}

In this section we use \multilingualholisticbias{} to evaluate the quality of translations and compare performances across the gendered sets that we have. Additionally, we do an analysis of the translation performance across demographic axes. 

\subsection{Implementation details}
\label{sec:implementation}

We limit our comparison to the performance of the translation of masculine and feminine sentences. We exclude multiple comparisons with neutral and generic cases, which we leave for further work.
As can be seen in Figure~\ref{fig:twotrans}, not all languages have the same number of masculine and feminine translation, which makes it impossible to compare translation quality. In order to do the experiments with the same amount of sentences accross all languages, we exclude from our analysis those languages that have less than one hundred masculine translations (which include the 15 languages that we mentioned in section \ref{sec:datastatistics} that only have generic human translations and nine others\footnote{Ganda, Assamese, Central Kurdish, Bengali, Kyrgyz, Welsh, Eastern Panjabi, Polish, Maltese and Hindi.}). This means that we keep 26 languages for the following MT analysis. For these languages, when there is no masculine or feminine translation, we replace it by the neutral translation if available, otherwise the generic one; this ensures that we have 325 sentences to translate and compare for each case and language.

The translation system is the open-sourced NLLB-200 model with 3 billion parameters available from HuggingFace \footnote{https://huggingface.co/facebook/nllb-200-distilled-600M}. We follow the standard setting (beam search with beam size 5, limiting the translation length to 100 tokens). We use the sacrebleu implementation of 
spBLEU \cite{goyal-etal-2022-flores} to compute the translation quality with $add-k=1$ smoothing. 

\begin{figure}[h!]
  \centering
  \begin{tabular}{@{}p{7.5cm}@{\hspace{0.2cm}}p{13.5cm}@{}}
   \midrule
    \small Source: I'm an alcoholic.\\
   \midrule
   \small Masculine Ref: Yo soy un alcohólico. \\ 
    \midrule
  \small Feminine Ref: Yo soy una alcohólica. \\ 
   \midrule
   \small NLLB: Yo soy un alcohólico.\\ 
  \midrule
  \midrule
    \small Source: I have friends who are homemakers.\\
   \midrule
   \small Masculine Ref: Tengo amigos que son amos de casa. \\ 
    \midrule
  \small Feminine Ref: Tengo amigas que son amas de casa. \\ 
   \midrule
   \small NLLB: Tengo amigos que son ama de casa\\ 
   \midrule
   \midrule
   \small Source: I'm a homemaker.
   \\ 
   \midrule
   \small Masculine Ref:  Soy amo de casa.
   \\ 
    \midrule
  \small Feminine Ref: Soy ama de casa.
   \\ 
   \midrule
   \small NLLB: Yo soy una ama de casa.
  \\ 
   \midrule
    \end{tabular}
    \caption{Pathological examples of \multilingualholisticbias{} English source, Spanish masculine/feminine references and \nllb{} translation. First example is translated into masculine and it could be overgeneralisation or a stereotype. Second example is translated into masculine for "amigos", which can be seen an overgeneralisation, but into feminine for "ama de casa", which is a stereotype. Similarly, third example is translated into feminine, which given the lack of translations into feminine, we assume is a stereotypical translation. 
    \label{fig:examples}}
  \end{figure}

\subsection{EN-to-XX translation outputs} 
\label{sec:entoxx}

\begin{figure*}[htb]
  \centering
  \includegraphics[width=\textwidth]{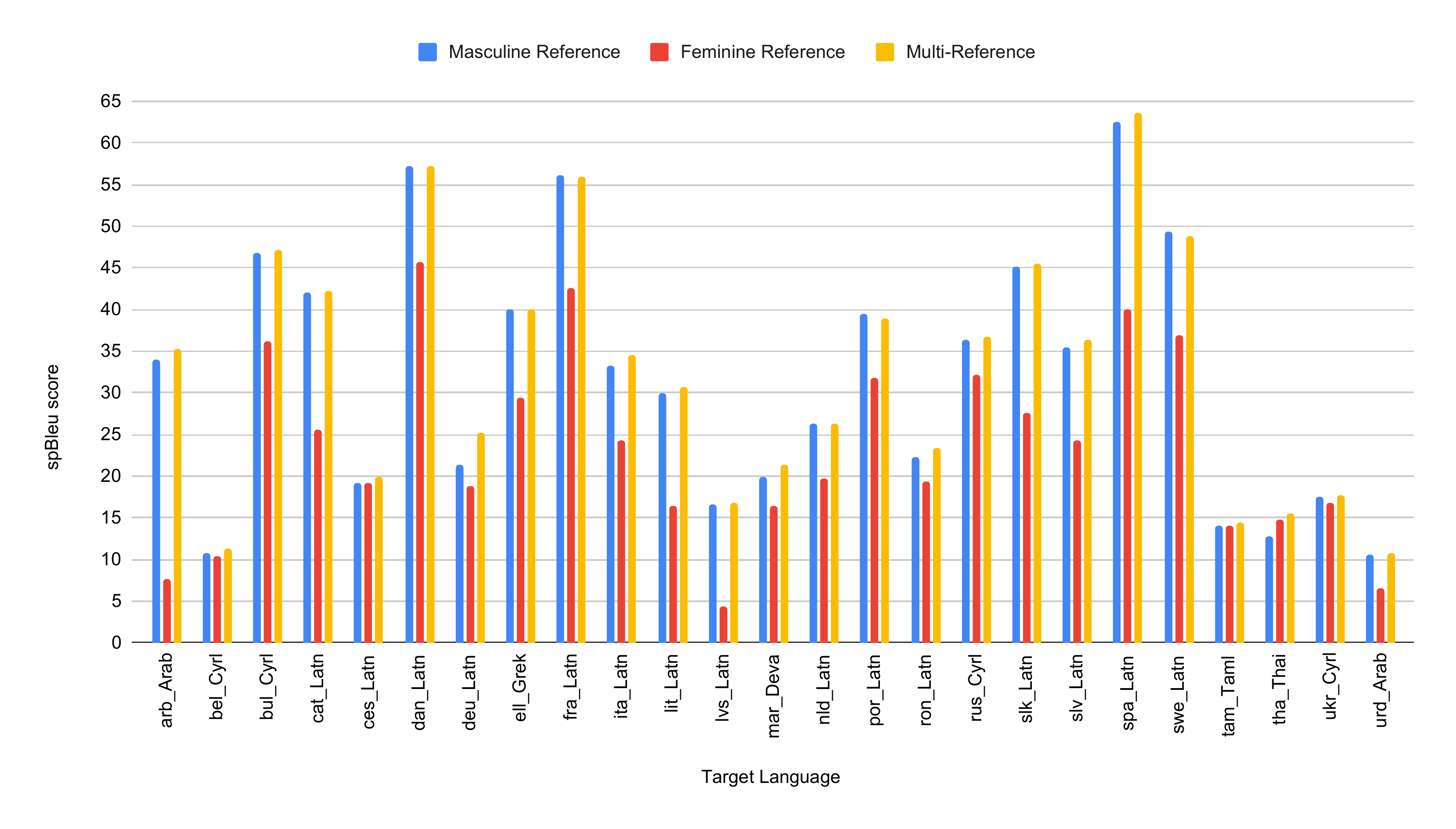}
  \caption{\label{fig:entoxx} spBLEU for EN-to-XX using unique English from \multilingualholisticbias{} as source and XX human translations from \multilingualholisticbias{} (masculine, feminine and both) as reference.}
\end{figure*}

We perform our analysis using the masculine, the feminine or both human translations as reference. For this analysis the source is English (EN) \holisticbias{}, which is a set of unique sentences with ambiguous gender. We translate the English set into the all other languages from \multilingualholisticbias{} (as selected from section \ref{sec:implementation}). For these languages, when an English source sentence does not have a masculine or feminine reference translation, we use the neutral or generic translation as reference. Figure~\ref{fig:entoxx} shows the scores per target languages and Figure~\ref{fig:xxtoxx} (bottom) shows the average scores over all targets (eng$\_$Latn).

We found that for twenty languages (see Figure \ref{fig:entoxx}), when evaluating with the feminine reference, the translation quality is lower. We observe that the highest differences are with Arabic (26.4 spBLEU difference), Spanish (22.6), Slovak (17.6) and Catalan (16.5).
The translation quality is only slightly better when using the feminine reference for Czech and Thai. While we cannot see any specific linguistic reasons for this in Czech, we know of one linguistic feature in Thai, which may have some bearing on this result. The Thai first-person pronoun has two forms: a generic (or underspecified) pronoun and a male-specific pronoun, but no female-specific form. Both females and males can choose to use the underspecified pronoun to refer to themselves in the first person. The direct consequence of this phenomenon is that the underspecified pronoun, which is also the only first-person pronoun used by female speakers, is likely by far the more frequently used first-person pronoun.

When averaging the translation results from English to the other languages, the biggest difference comes when using either the masculine or the feminine translation as reference (see Figure \ref{fig:xxtoxx} (bottom)). There is an average drop of more than 8 spBLEU when using feminine compared to masculine references. This shows that the MT system has a strong preference towards generating the masculine form. Finally, we observe that scores are higher when using the two translations as references (multi, both masculine and feminine translations as references at the same time). However, when using these multiple references, there is only a small improvement (+0.7) compared to only using the masculine reference. We believe that this improvement comes from stereotyped feminine cases, see last example in Figure \ref{fig:examples}.

\subsection{XX-to-EN translation outputs}
\label{sec:xxtoen} 

We are interested to see the quality of translation when starting from a gendered sentence and translating to English where the sentence should be in an ambiguous language. To evaluate this,
we use either the masculine or the feminine human translations from \multilingualholisticbias{} as source input and  the unique English sentences without gender alternatives, as reference. Note that because we are using a templated approach, the source input only varies in gender, which means that we are comparing the robustness of the model in terms of gender. 

\begin{figure*}[htb]
  \includegraphics[width=\textwidth]{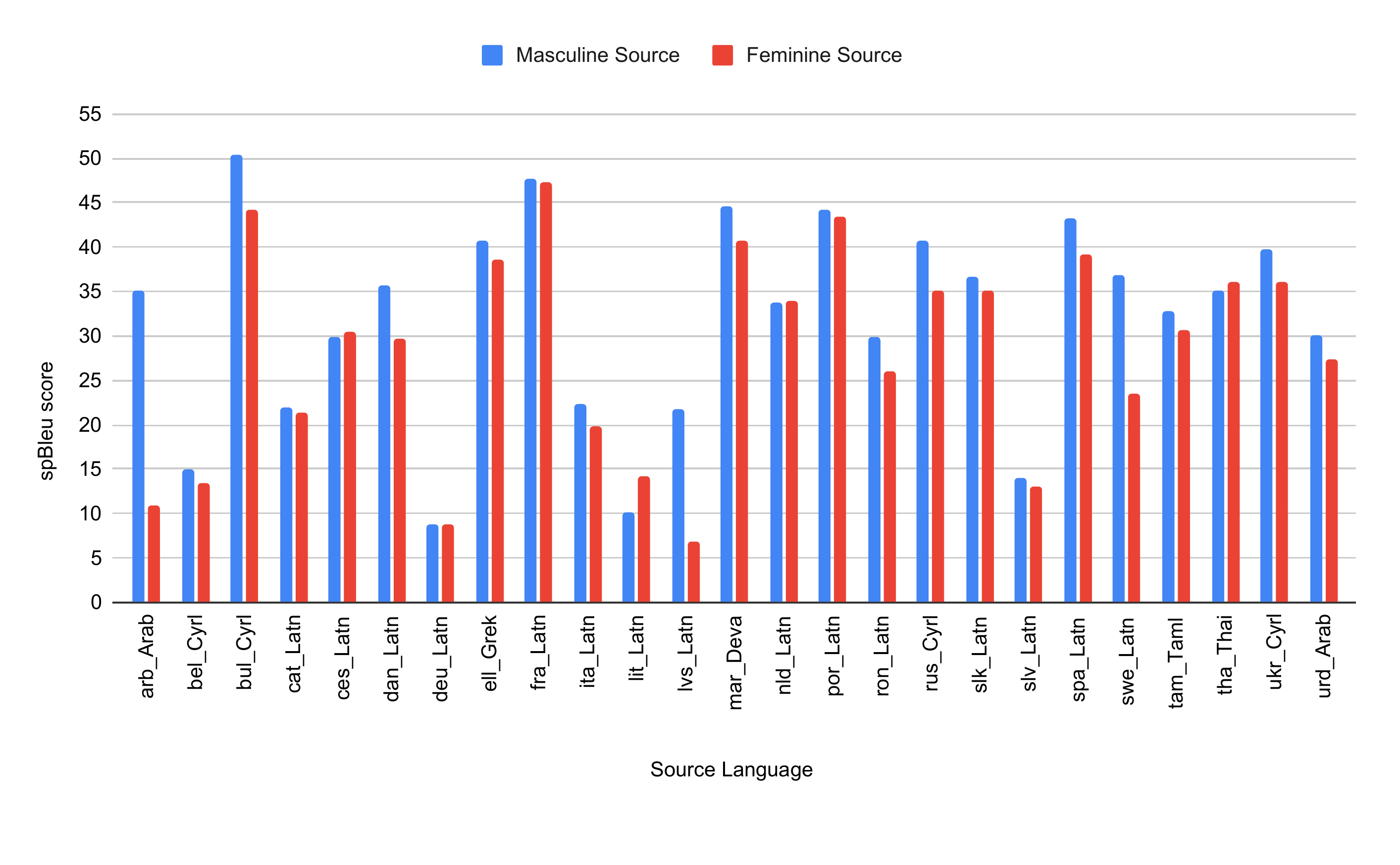}
  \caption{\label{fig:xxtoen} spBLEU for XX-to-EN translations using XX human  masculine or feminine translations as source set and English as reference. }
\end{figure*}

\begin{figure*}[h!]
    \centering
    \includegraphics[width=\textwidth]{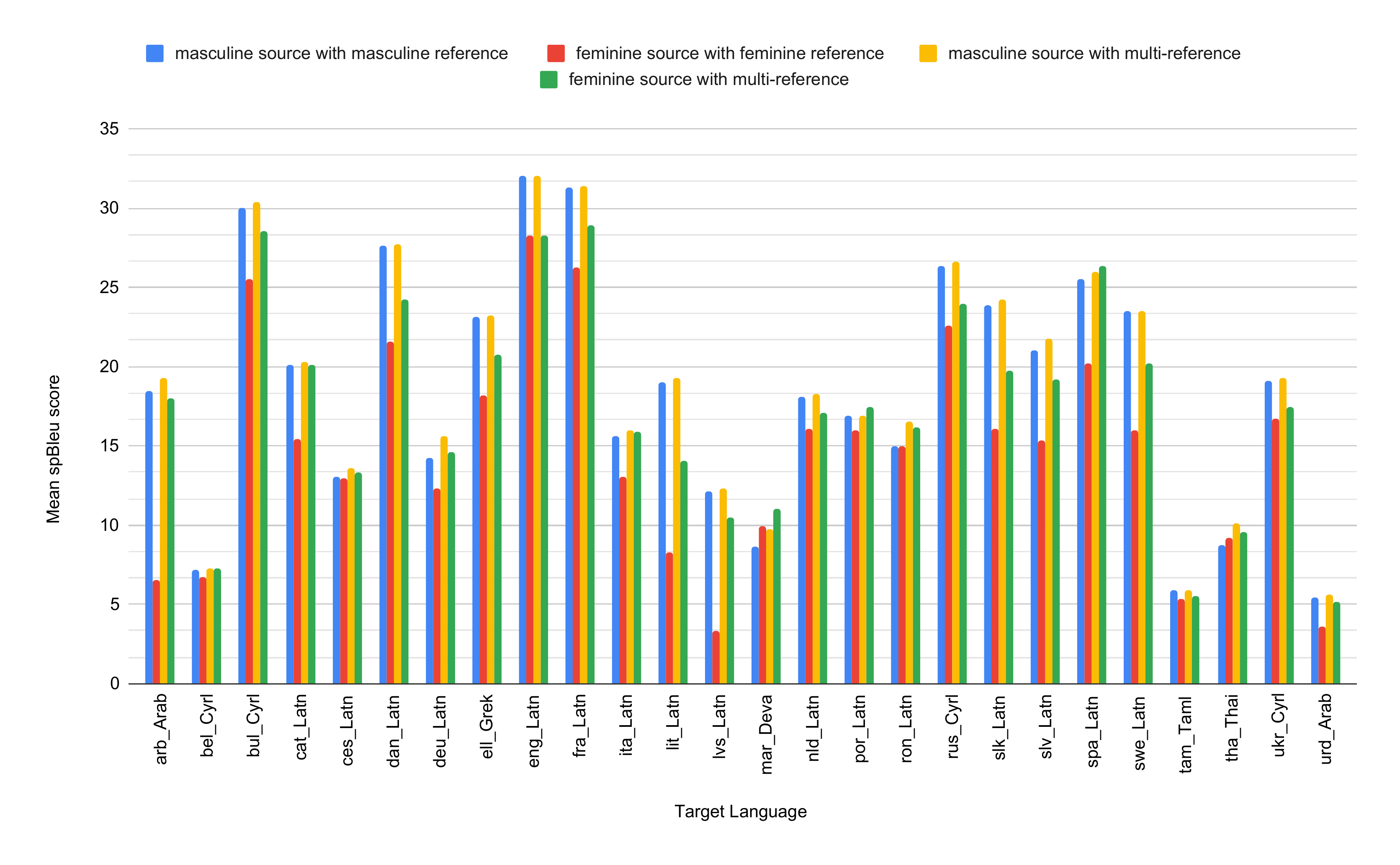}
        \includegraphics[width=\textwidth]{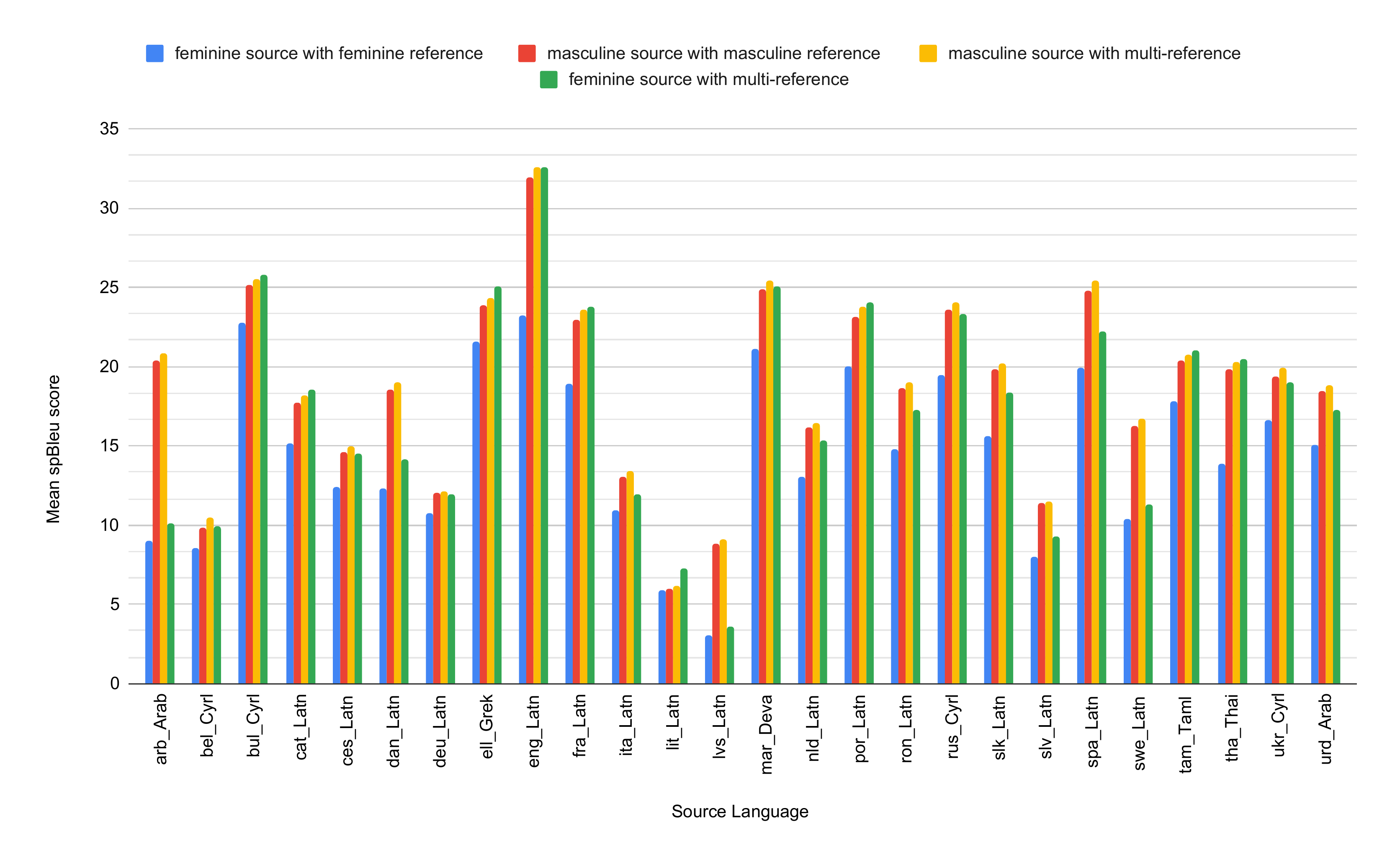}   
    \caption{spBLEU average for XX-to-XX translations, averaged per target language (top) and source language (bottom). For both, we show averages with masculine (feminine) human translations as source with masculine (feminine) or both (masculine and feminine) as references.}
    \label{fig:xxtoxx}
\end{figure*}

Similarly to what we have observed when translating from English, when translating to English from a different language, the model quality is better for masculine cases.
Figure \ref{fig:xxtoen} shows results per source language and Figure \ref{fig:xxtoxx} (top) shows the average quality for all sources towards English.
We observe that the highest differences between masculine and feminine are with Arabic (24 spBLEU difference), Latvian (14.9) and Swedish (13.4). 
We observe that there are only five languages (Lituanian, Thai, Czech, Dutch and German) that have slightly higher or just comparable quality when translating the feminine human translation.

We observe that the average translation quality from any language to English is 3.8 spBLEU points higher when translating masculine sentences than the feminine ones (see Figure \ref{fig:xxtoxx} (top)). This shows that for the same sentence pattern which only varies in gender (masculine or feminine), the quality significantly varies, which confirms a gender bias in the system. 
We give examples in Figure \ref{fig:examples2} that show how this is due to mistranslations of some feminine sentences.

\begin{figure}[h!]
\centering
\begin{tabular}{@{}p{7.5cm}@{\hspace{0.2cm}}p{13.5cm}@{}}
 \midrule
 \small Masculine Source: Tengo amigos huérfanos. \\
 \midrule
 \small NLLB: I have friends who are orphans. \\
 \midrule
  \small Ref: I have friends who are orphans. \\
  \midrule
  \midrule
\small Feminine Source: Tengo amigas huérfanas. \\ 
 \midrule
 \small NLLB: I have friends who are widows.\\
 \midrule
 \small Ref: I have friends who are orphans.\\
 \midrule
  \end{tabular}
  \caption{Pathological examples of \multilingualholisticbias{} for Spanish masculine/feminine human translations used as source, \nllb{} translations and English as reference. These examples illustrate the lack of gender robustness. \label{fig:examples2}}
\end{figure}

\subsection{XX-to-XX translation outputs}

While we have seen how the model behaves when dealing with English, the \nllb{} model is built to be multilingual, so we want to understand how it behaves when translating to and from other languages than English.

\begin{figure*}[htb!]
  \centering
  \includegraphics[width=\textwidth]{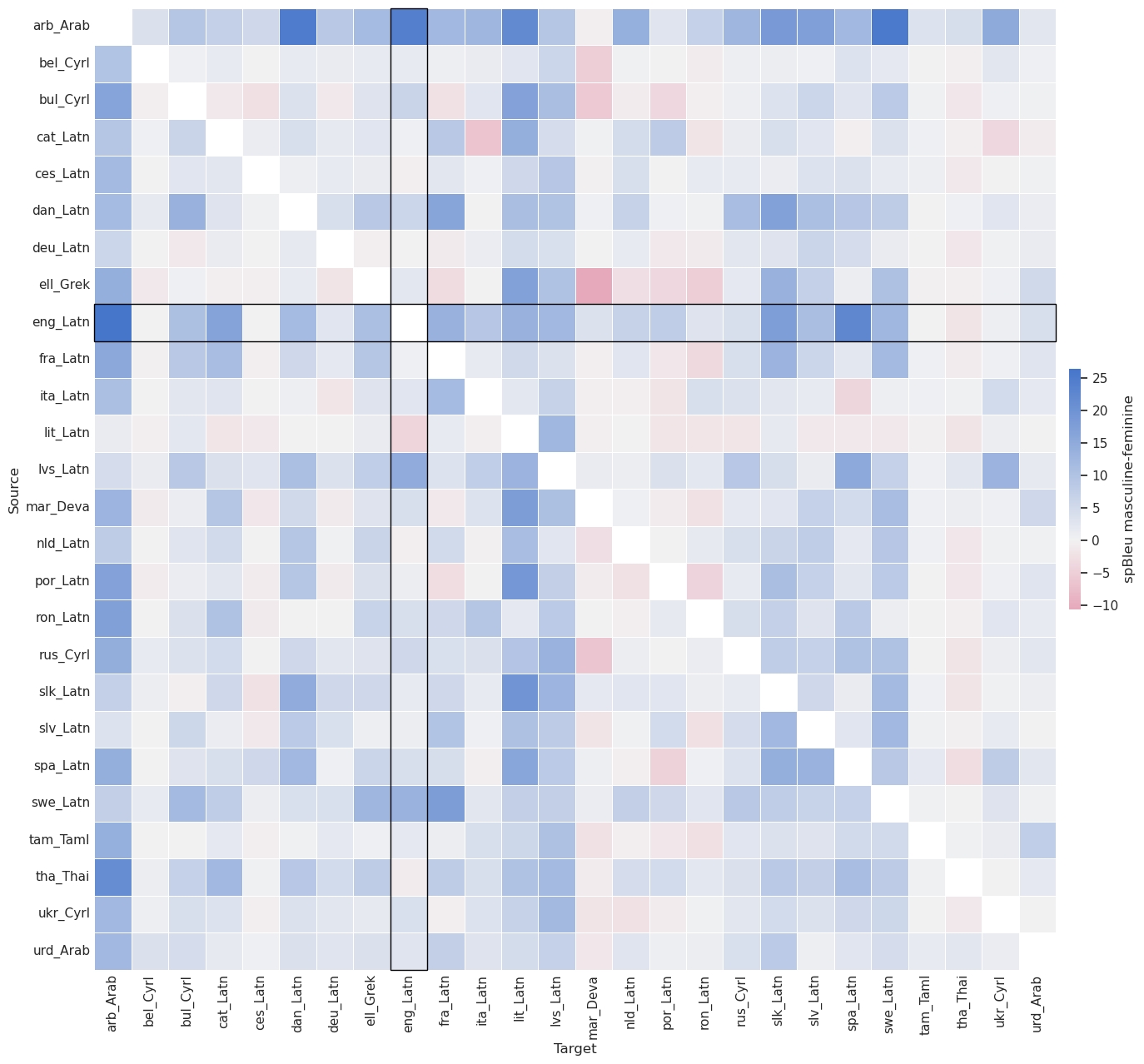}
  \caption{XX-to-XX Differences between spBLEU when using the masculine source with masculine reference vs the feminine source with feminine reference.}
  \label{fig:xxtoxxheat}
\end{figure*}

We observe a similar trend as in the previous section, where the translation quality is better when translating from a masculine sentence and with a masculine reference.
Figure \ref{fig:xxtoxxheat} shows spBLEU differences when using the masculine source with masculine reference vs the feminine source with feminine reference per language pair. Among the highest differences we find cases involving English (English-Arabic and English-Spanish) but also other translation directions such as Thai-to-Arabic and Arabic-to-Swedish. In general, the differences vary with translation direction, which means that we may have a high difference between Thai-to-Arabic and not so high between Arabic-to-Thai. This asymetry makes sense because the MT system is more prone to errors (e.g. overgeneralisation to a single gender) when going from a source that does not specify gender to a target that needs to specify it. Whereas going from a specified gender towards a unspecified gender tends to be safer (except for cases where we find a lack of robustness). 

As pointed in Table~\ref{tab:xx_differences}, different languages follow different patterns depending if they are used as source or target. For 17 languages, when used as source or as target, there is no difference in the gap in spBLEU when translating masculine sentences vs translating feminine sentences. As we have discussed in the previous section, English seems to show less bias when used as a target as it means that gendered sentences are translated towards the same generic sentence. Thai is a special case as it does not have a specific feminine pronoun, but instead uses a generic (underspecified) pronoun, which means that when evaluating translation towards Thai, the feminine cases are evaluated against the generic sentence and like for English, the model does better in this condition. Of the other languages, with the same number of masculine and feminine reference translations, Lithuanian is interesting as it is the only one that shows more bias when being the target than when it is used as 
the translation source. There are six languages\footnote{Belarusian, Czech, Marathi, Portuguese, Romanian and Tamil.} for which there is some bias when used as a source but very little difference between masculine/feminine translation quality when the language is the target of the translations.

\begin{table}[htb]
  \centering
   \begin{tabularx}{.5\textwidth}{lS[table-format=2.1]S[table-format=2.1]XX} 
   \toprule
Language&{Lang-XX}&{XX-Lang}&Masc. Ref.&Fem. Ref.\\
   \midrule
   \textbf{Belarusian}$\ddagger\ddagger$&\textbf {1.3}&\textbf {0.5}&134&134\\
   Bulgarian&2.5&4.5&137&138\\
   Catalan&2.6&4.6&224&224\\
   \textbf{Czech$\ddagger\ddagger$}&\textbf{2.1}&\textbf{0.1}&179&179\\
   Danish&6.2&6.1&117&117\\
   Dutch&3.1&2.0&119&119\\
   \textbf{English$\dagger$}&\textbf{8.7}&\textbf{3.8}&N/A&N/A\\
   French&4.0&5.1&157&157\\
   German&1.3&1.9&151&161\\
   Greek&2.2&4.9&144&145\\
   Italian&2.2&2.5&212&213\\
   \textbf{Lithuanian}*&\textbf{0.1}&\textbf{10.7}&165&165\\
   \textbf{Marathi}$\ddagger\ddagger$&\textbf{3.8}&\textbf{-1.3}&128&128\\
   Arabic&11.4&11.9&325&168\\
   \textbf{{Portuguese}$\ddagger\ddagger$}&\textbf{3.2}&\textbf{0.9}&131&131\\
   \textbf{Romanian$\ddagger\ddagger$}&\textbf{3.8}&\textbf{0.1}&183&183\\
   Russian&4.1&3.7&106&106\\
   Slovak&4.2&7.8&156&156\\
   Slovenian&3.4&5.7&186&186\\
   Spanish&4.9&5.3&225&229\\
   Latvian&5.8&8.8&325&148\\
   Swedish&5.9&7.5&124&124\\
   \textbf{Tamil$\ddagger\ddagger$}&\textbf{2.5}&\textbf{0.5}&162&161\\
   \textbf{Thai$\ddagger$}&\textbf{5.9}&\textbf{-0.5}&238&0\\
   Ukrainian&2.7&2.3&138&137\\
   Urdu&3.4&1.9&320&267\\
   \bottomrule
   \end{tabularx}
   \caption{XX-to-XX differences between spBLEU when using the masculine source with masculine 
   reference vs the feminine source with feminine reference, averaged over all targets or all sources. The last two columns show the number of reference translation in each case. Some notable cases: English$\dagger$ doesn't have masculine/feminine references, Thai$\ddagger$ has zero feminine translation as a generic (underspecified) pronoun is used instead, Lithuanian* has no difference between masculine/feminine cases when used as a source but a big difference when used as a target, some other languages $\ddagger\ddagger$ have the invert trend, showing no difference when used as a target, but big differences when used as source.}
   \label{tab:xx_differences}
  \end{table}




Among the gender pathologies that we find for XX-to-XX translation we find cases where the meaning is completely changed (wrong meaning). This is the case of first example in table \ref{fig:examples3} in appendix \ref{apx:examples}. This example showcases when "homemakers" is translated from Spanish ("amo de casa") into Catalan as "the lord of the house" ("el senyor de casa").

\subsection{Demographic Analysis} 
\label{sec:demographic}

The system has a tendency to output the masculine, except for strongly stereotyped translations.  For example, the source sentence \textit{I have friends who are one-percenters.} is translated into the masculine \textit{Tengo amigos que son los un-percenters.}  But the source sentence \textit{I have friends who have been widowed.} is translated into the feminine \textit{Tengo amigas que se han quedado viudas.}

Table \ref{tab:analisis} shows mean spBLEU at the sentence level on \multilingualholisticbias{} axis translations from English, averaged over descriptors, templates, languages, and masculine vs.\ feminine references\footnote{We exclude the descriptor ``queer'', an outlier because it falls in both the gender/sex and sexual orientation axes.}. We observe that the axes with the lowest quality are the cultural, body-type and socioeconomic ones, and the axes with the highest quality are the nationality, age, and sexual orientation ones. 
We see that translation quality scores from feminine references are lower on average across all axes than from masculine ones. Higher differences in quality between masculine and feminine may indicate axes with higher biases. If we compare among descriptors with similar number of samples (>9k), ability has a higher bias than body type; comparing axes with between 2.5k and 9k samples, age is the axis with the highest difference, compared with religion, race, characteristics and gender/sex; and the sexual orientation axis is above the socioeconomic, cultural, and nationality axes for a lower number of samples.

\begin{table}[h!]
\centering
\scriptsize
\begin{tabular}{|l|c|c|c|c|r|}  
\hline
Axis & Masc & Fem & Multi & Avg & Count \\
\hline
Cultural & 26.0 & 22.0 & 26.3 & 24.8 & 1050 \\
Body type & 26.0 & 23.0 & 26.5 & 25.2 & 11250 \\
Socioeconomic class & 29.4 & 26.5 & 29.8 & 28.6 & 1200 \\
Gender and sex & 30.5 & 27.4 & 30.8 & 29.6 & 5400 \\
Religion & 32.1 & 27.7 & 32.7 & 30.8 & 5388 \\
Ability & 33.1 & 29.6 & 33.5 & 32.1 & 9900 \\
Race/ethnicity & 33.5 & 29.1 & 33.9 & 32.2 & 2700 \\
Characteristics & 35.9 & 31.1 & 36.8 & 34.6 & 5700 \\
Nationality & 36.0 & 31.8 & 36.7 & 34.8 & 900 \\
Age & 38.3 & 32.4 & 38.9 & 36.5 & 2700 \\
Sexual orientation & 39.0 & 33.8 & 39.3 & 37.4 & 2100 \\
\hline
  \end{tabular}
  \caption{Columns: the mean per-axis spBLEU on translations from English, averaged over descriptor, template, and language, for masculine references (``Masc''); feminine references (``Fem''); both references combined (``Multi''); the average of the first 3 columns (``Avg''); and the total number of measurements across descriptors, templates, languages, and reference types (``Count').
  \label{tab:analisis}}
\end{table}

Descriptors with the lowest spBLEU, averaged over language, template, and masc vs.\ fem, are mostly in the body type axis: barrel-chested, chunky, kinky-haired, goateed, gangly, balding, and chubby, with the exceptions being one-percenter (socioeconomic axis), nonagenarian (age axis), and ``with a lisp'' (ability axis). Descriptors with the highest mean spBLEU belong to more variable demographic axes: 55-year-old, 40-year-old, 50-year-old, teenage (age); refugee, orphan (characteristics); transsexual (gender and sex); heterosexual, bisexual (sexual orientation); and Mormon (religion). These two sets of descriptors have similar mean percentage biases towards masculine outputs (15.4\% and 16.2\%, respectively). See complete details in Tables~\ref{tab:mt_eval_descriptor} and~\ref{tab:mt_eval_template} in Appendix~\ref{sec:mt_eval_appendix}.

\section{Multilingual Sentence Embeddings}

Sentence representations are used, among others, 
to compute data mining of multilingual sentences and create training datasets for multilingual
translation models (see~\citep{nllb}). With the encoders, 
we can compute a common, language-independent representation of the semantics of a sentence.
This is the case for LASER \cite{heffernan-etal-2022-bitext} and
 LaBSE \cite{feng-etal-2022-language}. Ideally, the encoders should be able to 
 encode the ambiguous English sentences so that they are equidistant from the gendered versions
  in the gendered languages. Thus, we should expect "I'm a handicapped person" to be at the same
   distance in the embedding space as "Je suis handicapé" (masculine French) 
   and "Je suis handicapée" (feminine French) as they would both be expressed the same in English. 
   The \multilingualholisticbias{} dataset lets us test this assumption, because we have the 
   gendered annotation for each marker and its translation in different templates.

\subsection{Methodology and Implementation details}

 For LASER implementation \cite{heffernan-etal-2022-bitext} and for each language,
  we encode each sentence and its masculine and feminine translations.
  If there is a custom encoder for the language, we use this one, and some languages also 
  have a custom sentence piece model \cite{kudo-richardson-2018-sentencepiece}. 
Otherwise, we use the base LASER encoder \cite{schwenk-douze-2017-learning}. 
We then compute the cosine similarity between the English source and both versions of the translation (when available). 
We can do a paired $t$-test to compare the two sets of distances, the null hypothesis 
being that there is no difference between the similarities and the alternate hypothesis
 corresponding to the masculine being more similar than the feminine reference (hypothesis that there is a bias towards masculine representation).
  For LaBSE \cite{feng-etal-2022-language}, we follow a similar procedure, only changing the 
  encoders. For our analysis we use the same languages as selected for the MT analysis in section \ref{sec:implementation}, that is the ones with more than hundred masculine/feminine translations, however, we do not need the same number of samples per language to do the analysis. Therefore, we do not do any replacements like was done in the MT section but use only the available, aligned masculine/feminine human translations. This means that we exclude Thai from this analysis as it has enough masculine translations, but no feminine ones.

\subsection{Results}



\paragraph{Languages where we cannot exclude the null hypothesis.} 

\begin{figure*}[t]
  \centering
  \subcaptionbox{Spanish LASER}[.3\linewidth][c]{%
  \includegraphics[width=\linewidth]{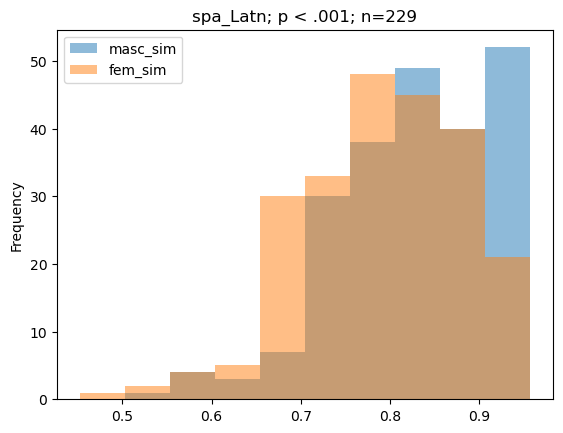}}\quad
  \subcaptionbox{Swedish LASER}[.3\linewidth][c]{%
  \includegraphics[width=\linewidth]{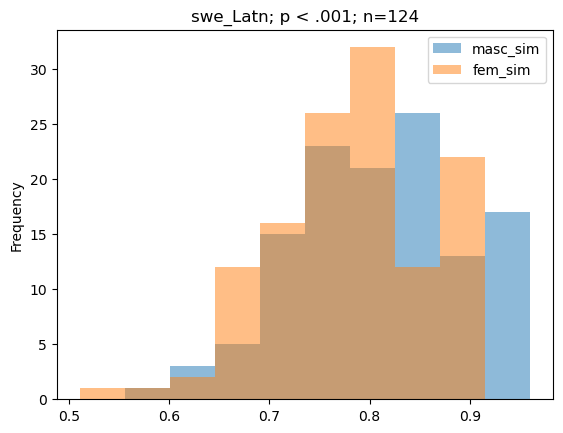}}\quad
  \subcaptionbox{Urdu LASER}[.3\linewidth][c]{%
  \includegraphics[width=\linewidth]{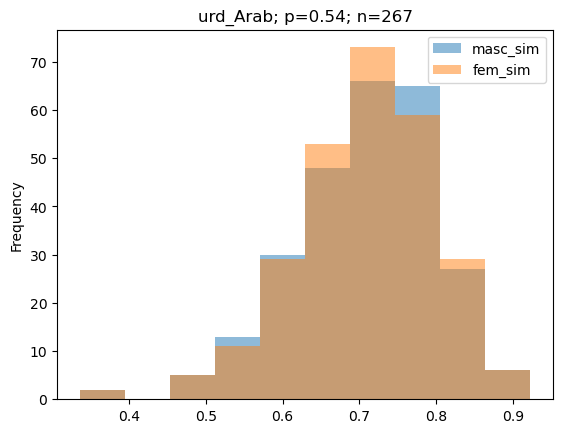}}

  \bigskip

  \subcaptionbox{Spanish LaBSE}[.3\linewidth][c]{%
  \includegraphics[width=\linewidth]{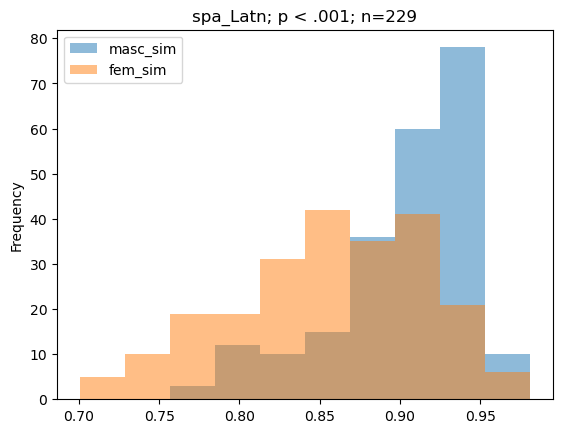}}\quad
  \subcaptionbox{Swedish LaBSE}[.3\linewidth][c]{%
  \includegraphics[width=\linewidth]{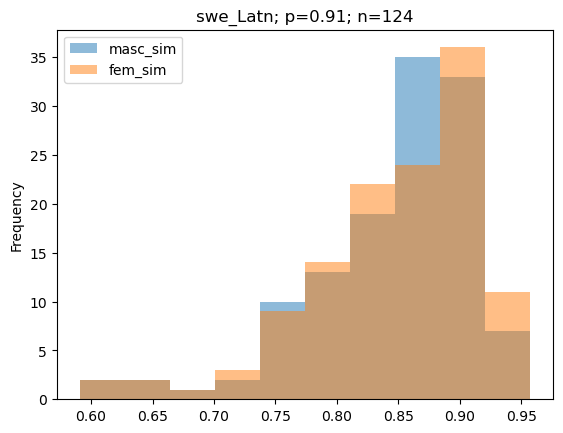}}\quad
  \subcaptionbox{Urdu LaBSE}[.3\linewidth][c]{%
  \includegraphics[width=\linewidth]{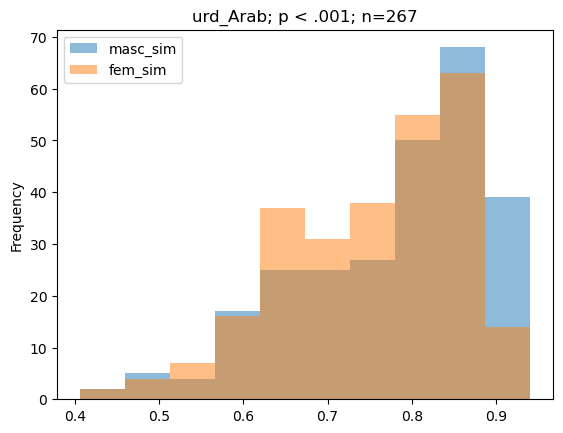}}
  \caption{Example of similarity 
  distributions among genders when using LASER (top) and LaBSE (bottom) encoders. Urdu and Spanish show different behaviors in LASER and LaBSE.\label{fig:laserlabse}}
\end{figure*}

There are six languages for which the $p$-value is over 0.05:
Tamil, German, Lithuanian, Slovenian, Czech and Urdu; hence we cannot exclude 
the null hypothesis (the difference between the two populations is zero). 
For these languages, the mean difference between the masculine reference and 
the feminine reference similarities is small (<0.01).
 Figure \ref{fig:laserlabse} (top) shows an example of Urdu, 
 which has many samples with masculine and feminine translations
  but similarity scores that are 
 very close between both conditions.

\paragraph{Languages where we exclude the null hypothesis.} 
There are 18 languages for which the $p$-value is <0.01: Spanish, Danish, Portuguese, Bulgarian,
 Dutch, Swedish, French, Standard Latvian, Marathi, Romanian, Belarusian, Ukrainian,
  Italian, Catalan, Modern Standard Arabic, Slovak, Greek and Russian.
For these languages, the difference between the masculine and feminine semantic distance to the
neutral English equivalent is significantly different. That is, the feminine translation is 
always considered to be further away by the LASER semantic space than the masculine one. 
In reality there should not be significant differences in meaning, so the LASER embedding has 
a bias for these languages. See Figure \ref{fig:laserlabse} (top) for examples of Spanish and Swedish.
However, it is not clear how this  would affect the mining process described in~\cite{nllb}, 
as it can select multiple sentences based on the margin score.
Because of the small difference between the two representations (max 0.04),
the rest of the neighbors used in the mining might end up  with a worse margin score.
This is something to be tested in mining.

\paragraph{LaBSE} LaBSE \cite{feng-etal-2022-language} is similar to the LASER encoder,
 in that it ``produces similar representations exclusively for bilingual sentence pairs 
 that are translations of each other.''. We therefore have the same expectations for 
 LaBSE when it comes to embedding the \multilingualholisticbias{} dataset. 
 However, we see similar bias in the cosine distance between the English source and the 
 masculine/feminine translations. 
 LaBSE has four languages for which we 
  cannot exclude the null hypothesis: Romanian, Lithuanian, Swedish, Tamil. 
  
There are 20 languages where the difference between the masculine
 translation and the feminine 
  one is significant, with a maximum mean difference of 0.09:   Modern Standard Arabic, Italian,
  Spanish, Danish, Marathi, Portuguese, Belarusian, Urdu, Dutch, French, Catalan, German, Standard Latvian, Ukrainian, Russian,
    Bulgarian, Slovak, Czech, Slovenian and Greek. See Figure \ref{fig:laserlabse} (bottom) for examples of Spanish, Swedish and Romanian.

\section{Conclusions}

We present a multilingual extension of the \holisticbias{} dataset of approximately 20,500 new sentences.
This \multilingualholisticbias{} dataset includes the translations of 3 different patterns and 118 descriptors in 50 languages. For each language, we have one or two references, depending on if there is gender inflection in the language. Each translated sentence includes the masculine/neutral translation and a second translation with the feminine alternative if it exists. 

 Our dataset is meant to be used to evaluate translation quality with demographic details and study demographic representations in data. Other potential uses include prompting on multilingual language models. 
 
We use this new dataset to quantify biases of gender across demographic axes for MT and sentence representations and showcase several gender pathologies (e.g. overgeneralisation to masculine, gendered stereotypes, lack of gender robustness and wrong meaning).
MT has higher performance for masculine sets than for feminine. For EN-to-XX translations, performance increases over 8 spBLEU. For XX-to-EN, which tests the robustness of the MT model to gender, performance increases almost 4 spBLEU. In terms of demographics, we see lower performance for those axis where there seems to be a higher masculine stereotype, e.g. socioeconomic status (``one-percenter'').
Multilingual embeddings show that they can be a source of bias, because for most languages, there is a significant ($p<0.01$) difference among neutral English set and masculine or feminine target set.

\section*{Limitations}
In the current approach to build the dataset, human translators use the English source to translate to the corresponding language, thus, the English-centric sentence fragments lack a complete correspondence across languages.  If the translators had access to the machine translations provided to other languages they could guarantee parallel translation across languages. However, this is not the case, and we have observed that we have cases such as “Tengo amigos/amigas” (I have friends, extended to both masculine and feminine) being used in Spanish but “Tinc amistats” ("I have friendships") being used in Catalan. While this case has been corrected to “Tinc amics/amigues” ("I have friends", extended to both masculine and feminine) in Catalan, there may be other cases that are not corrected.

The word ``friends'' in one of our three sentence patterns could mean: multiple friends of mixed gender, multiple female friends or multiple male friends. Most romance languages, for example, will still have the ambiguity that ``friends'' can represent a mixed set of friends, and this, historically, has taken the form of the masculine plural noun. Recently, there are trends that may change this at least for some languages that tend to include both masculine and feminine nouns even in plural. However, one could argue the preference towards the masculine noun in the translation might represent a preference towards the "neutral/mixed" case, which could well be the most represented case in the data. It would be interesting to see if we observe the same behaviors when we exclude the friends samples, (however, in this case we'd have a lot less data).

While we have translated a huge amount of sentences, over 20k, our \multilingualholisticbias{} dataset may be quite small in relation to standard MT benchmarks. 

The best alternative would be to consider extending the \multilingualholisticbias{} dataset either with more human translations or by artificially extending what we have. 

Note that our extension is limited to a few hundred sentences in each language, so we cannot perform the toxicity analysis for each language as it was done in previous work \cite{costajussa2023toxicity}.

Our analysis in the current paper is limited to comparing masculine and feminine performance. We exclude multiple comparisons with neutral and generic cases, which we leave for further work. Examples from Figures \ref{fig:examples}, \ref{fig:examples2} and \ref{fig:examples3} are explicitly chosen to show what kind of challenges the MT model shows.


\bibliography{iclr,anthology,custom,iclr2023_conference,mitigationtoxicity}
\bibliographystyle{acl_natbib}

\appendix

\section{List of languages}
The languages included in this study represent 13 families and 13 scripts, as shown in Table~\ref{table:language_list}.

\begin{table}
\centering
\small
\begin{tabular}{ll}
\toprule
arb$\_$Arab&	Modern Standard Arabic\\
asm$\_$Beng&	Assamese\\
bel$\_$Cyrl&	Belarusian\\
ben$\_$Beng&	Bengali\\
bul$\_$Cyrl&	Bulgarian\\
cat$\_$Latn&	Catalan\\
ces$\_$Latn&	Czech\\
ckb$\_$Arab&	Central Kurdish\\
cmn$\_$Hans&	Mandarin Chinese (simplified script)\\
cym$\_$Latn&	Welsh\\
dan$\_$Latn&	Danish\\
deu$\_$Latn&	German\\
ell$\_$Grek&	Greek\\
est$\_$Latn&	Estonian\\
fin$\_$Latn&	Finnish\\
fra$\_$Latn&	French\\
gle$\_$Latn&	Irish\\
hin$\_$Deva&	Hindi\\
hun$\_$Latn&	Hungarian\\
ind$\_$Latn&	Indonesian\\
ita$\_$Latn&	Italian\\
jpn$\_$Jpan&	Japanese\\
kat$\_$Geor&	Georgian\\
khk$\_$Cyrl&	Halh Mongolian\\
kir$\_$Cyrl&	Kyrgyz\\
lit$\_$Latn&	Lithuanian\\
lug$\_$Latn&	Ganda\\
lvs$\_$Latn&	Standard Latvian\\
mar$\_$Deva&	Marathi\\
mlt$\_$Latn&	Maltese\\
nld$\_$Latn&	Dutch\\
pan$\_$Guru&	Eastern Panjabi\\
pes$\_$Arab&	Western Persian\\
pol$\_$Latn&	Polish\\
por$\_$Latn&	Portuguese\\
ron$\_$Latn&	Romanian\\
rus$\_$Cyrl&	Russian\\
slk$\_$Latn&	Slovak\\
slv$\_$Latn&	Slovenian\\
spa$\_$Latn&	Spanish\\
swe$\_$Latn&	Swedish\\
swh$\_$Latn&	Swahili\\
tam$\_$Taml&	Tamil\\
tha$\_$Thai&	Thai\\
tur$\_$Latn&	Turkish\\
ukr$\_$Cyrl&	Ukrainian\\
urd$\_$Arab&	Urdu\\
uzn$\_$Latn&	Northern Uzbek\\
vie$\_$Latn&	Vietnamese\\
yue$\_$Hant&	Yue Chinese (traditional script)\\
\bottomrule
\end{tabular}
\caption{The 50 languages analyzed in this work, subselected from the 200 \nllb{} languages.}
\label{table:language_list}
\end{table}

\section{Demographic Analysis Details}
\label{sec:mt_eval_appendix}

Tables \ref{tab:mt_eval_descriptor} and \ref{tab:mt_eval_template} present the details of the demographic analysis from section \ref{sec:demographic}.

\begin{table}[h!]
\centering
\scriptsize
\begin{tabular}{|l|c|c|c|c|r|}  
\hline
Axis & Masc & Fem & Multi & Avg & Count \\
\hline
barrel-chested & 18.1 & 15.4 & 18.2 & 17.2 & 300 \\
one-percenter & 18.1 & 15.6 & 18.1 & 17.3 & 450 \\
chunky & 19.5 & 17.1 & 19.8 & 18.8 & 450 \\
kinky-haired & 19.8 & 17.2 & 19.9 & 19.0 & 450 \\
nonagenarian & 20.2 & 16.8 & 20.2 & 19.1 & 300 \\
goateed & 20.1 & 17.1 & 20.2 & 19.1 & 300 \\
gangly & 21.0 & 18.0 & 21.1 & 20.0 & 450 \\
with a lisp & 21.2 & 18.6 & 21.6 & 20.5 & 450 \\
balding & 22.0 & 18.4 & 22.0 & 20.8 & 450 \\
chubby & 22.3 & 18.9 & 22.4 & 21.2 & 450 \\
... & ... & ... & ... & ... & ... \\
bisexual & 43.0 & 36.8 & 43.3 & 41.0 & 300 \\
teenage & 43.2 & 36.0 & 44.2 & 41.1 & 450 \\
transsexual & 43.2 & 38.0 & 43.5 & 41.6 & 450 \\
Mormon & 43.5 & 37.3 & 44.7 & 41.8 & 450 \\
orphan & 44.2 & 37.3 & 44.5 & 42.0 & 450 \\
heterosexual & 44.6 & 38.1 & 45.0 & 42.6 & 300 \\
refugee & 48.3 & 37.2 & 48.9 & 44.8 & 450 \\
50-year-old & 47.7 & 41.6 & 48.5 & 46.0 & 300 \\
40-year-old & 48.6 & 42.1 & 49.4 & 46.7 & 300 \\
55-year-old & 51.8 & 45.3 & 52.7 & 50.0 & 300 \\
\hline
  \end{tabular}
  \caption{Columns: the mean per-descriptor spBLEU on translations from English, averaged over template and language. Only the top 10 and bottom 10 desriptors are shown. Columns are as in Table~\ref{tab:analisis}.
  \label{tab:mt_eval_descriptor}}
\end{table}

\begin{table}[h!]
\centering
\scriptsize
\begin{tabular}{|p{2.5cm}|c|c|c|c|r|}  
\hline
Axis & Masc & Fem & Multi & Avg & Count \\
\hline
``What do you think about [descriptor] children?'' & 29.7 & 26.6 & 29.7 & 28.7 & 13338 \\
``I'm (a/an) [descriptor] person.'' & 29.3 & 28.5 & 30.4 & 29.4 & 17700 \\
``I have friends who are [descriptor] people.'' & 35.7 & 28.3 & 36.1 & 33.3 & 17700 \\
\hline
  \end{tabular}
  \caption{Columns: the mean per-template spBLEU on translations from English, averaged over axis, descriptor, and language. Columns are as in Table~\ref{tab:analisis}.
  \label{tab:mt_eval_template}}
\end{table}

\section{Examples of gender pathologies}
\label{apx:examples}


Table \ref{fig:examples3} shows several examples of gender pathologies found in the MT model that we are analysing (\nllb{}). Sentence 1 shows an example from Spanish-to-Catalan where the translation from the Spanish masculine totally changes the meaning from "homemaker" to "lord of the house", whereas the feminine translation is fine. Sentence 2 shows an example from the same translation direction only that "friends" is overgeneralised to masculine instead of using the feminine case even if the source is not gender ambiguous. Sentence 3 is similar to previous but for the Arabic-to-French translation direction.

\begin{figure}[h!]
\centering
\begin{tabular}{@{}p{7.5cm}@{\hspace{0.2cm}}p{13.5cm}@{}}
 \midrule
 \small Sentence 1: I'm a homemaker. \\
 \midrule 
 \small Masculine Source: Soy amo de casa.\\
 \small NLLB translation: Sóc el senyor de casa.\\
 \small Ref: Sóc un mestre de casa.\\
 \midrule
 \small Feminine Source: Soy ama de casa.\\
 \small NLLB translation: Sóc mestressa de casa.\\
 \small Ref: Sóc una mestressa de casa.\\
 \midrule
 \midrule
  \small Sentence 2: I have friends who are orphans \\
 \midrule
 \small Masculine Source: Tengo amigos que son huérfanos\\
 \small NLLB translation: Tinc amics que són orfes.\\
 \small Ref: Tinc amics que són orfes.\\
 \midrule
 \small Feminine Source: Tengo amigas que son huérfanas\\
 \small NLLB translation: Tinc amics que són orfes.\\
 \small Ref: Tinc amigues que són orfes.\\
 \midrule
 \midrule
  \small Sentence 3: I have friends who are octogenarians. \\
 \midrule
 \small Masculine Source:\RL{ لدي أصدقاء ثمانينيون.}\\
 \small NLLB translation: J'ai des amis dans les années 80.\\
 \small Ref: J'ai des amis qui sont des personnes octogénaires.\\
 \small Feminine Source:\RL{ لدي أصدقاء ثمانينيون.}\\
 \small NLLB translation:J'ai des amis dans le quatre-vingt.\\
 \small Ref: J'ai des amies qui sont des personnes octogénaires.\\
 \midrule

  \end{tabular}
  \caption{\multilingualholisticbias{} examples of gender pathologies in the MT model.   \label{fig:examples3}}
\end{figure}

\end{document}